\documentclass{article}

\usepackage{arxiv}

\usepackage[utf8]{inputenc} 
\usepackage[T1]{fontenc}    
\usepackage{hyperref}       
\usepackage{url}            
\usepackage{booktabs}       
\usepackage{amsfonts}       
\usepackage{nicefrac}       
\usepackage{microtype}      
\usepackage{lipsum}
\usepackage{graphicx}
\graphicspath{ {./images/} }
\usepackage{tcolorbox}
\definecolor{myboxcolor}{RGB}{245,245,245} 
\definecolor{myframe}{RGB}{0,0,128} 
\usepackage{amsmath}
\newtcolorbox{mybody}{
  colback=myboxcolor,
  colframe=myframe,
  boxrule=1pt, 
  left=1pt,
  right=1pt,
  top=1pt,
  bottom=1pt,
}

\title{Zero Shot Coordination for Sparse Reward Tasks with Diverse Reward Shapings}

\author{
  Keenan Powell \\
  Department of Computer Science\\
  University of Maryland, College Park \\
  \texttt{kpowell1@terpmail.umd.edu} \\
   \And
  Peihong Yu \\
  Department of Computer Science\\
  University of Maryland, College Park \\
  \And
  Pratap Tokekar \\
  Department of Computer Science\\
  University of Maryland, College Park \\
  \texttt{tokekar@umd.edu} \\
}

\begin{document}
\maketitle
\begin{abstract}
  Many Multi-Agent Reinforcement Learning (MARL) agents fail to adapt properly to cooperating with agents trained with the same objectives but different seeds, algorithms, or other training differences. This is the problem of Zero-Shot Coordination (ZSC), which focuses on training agents to cooperate well with unknown agents. ZSC has been studied for a variety of tabular cases and simple games such as Hanabi, achieving excellent results. However, existing solutions to ZSC only consider identical rewards for your trained agents and all future partners. This is not realistic for the trained agents, as they do not consider the problem of cooperating with agents that have identical sparse objectives but shape the rewards for those objectives in different manner. To address this issue, we show how to train an ensemble of methods using randomized reward shapings chosen using 4 selection algorithms. Experiments done on the Overcooked environment demonstrate consistent improvements of 62.2\%-119.2\% in sparse reward over baseline ZSC algorithms when playing with agents that have identical sparse rewards but different reward shapings. 
\end{abstract}

\section{INTRODUCTION AND RELATED WORK}

Multi-Agent Reinforcement Learning (MARL) has emerged as a powerful tool for training agents capable of coordination across fields such as vehicle routing, self-driving vehicles, cooperative games, and multi-drone pursuit  \cite{liu2023learningplaytrajectorygames, foerster2019bayesianactiondecoderdeep, NEURIPS2020_65cf25ef, li2025atdronebenchmarkingadaptiveteaming, hu2021otherplayzeroshotcoordination}. However, traditional MARL algorithms and techniques often fail when attempting to cooperate with unknown partners. Because multiple optimal strategies can exist simultaneously under Self-Play \cite{10.1162/neco.1994.6.2.215}, agents can easily form conventions and pre-coordinated strategies that don't work when attempting to coordinate with agents that have formed different conventions and strategies. Zero-Shot Coordination (ZSC) addresses this issue by training agents to form a policy that can adapt to a wide variety of other partners (in this paper, other trained agents) without requiring additional training with the new partner. While many existing ZSC approaches have achieved excellent results, including Other-Play \cite{hu2021otherplayzeroshotcoordination}, which takes advantage of symmetries in the environment, TrajeDi \cite{pmlr-v139-lupu21a}, which uses ensemble learning with a diversity component in the loss, and MAZE \cite{xue2024heterogeneousmultiagentzeroshotcoordination}, which co-evolves heterogenous populations, these approaches generally either assume or rely upon identical reward functions between different agents. This is unrealistic: while cooperating agents will likely share sparse objectives, their reward shapings (weights for environmental features to densify a sparse reward) \cite{ibrahim2024comprehensiveoverviewrewardengineering}, will often vary depending on where and how they were trained. For example, self-driving cars from multiple companies may share a sparse objective of reaching a destination but weight aspects like speed vs. safety differently. Hidden-Utility Self-Play (HSP) \cite{yu2023learningzeroshotcooperationhumans} is likely the most similar algorithm to ours, extending the standard self-play framework for zero-shot cooperation by explicitly modeling human biases as hidden reward functions. HSP generates a diverse policy pool by producing agents with varied biases through different features. While HSP can produce different rewards, it uses a different method than ours (focusing on a diverse policy pool for a final agent to play against instead of a ensemble method), and it's meant for specific human biases based on an existing environmental reward. It also relies on purely random feature weighting, without any enforcement of diversity or performance. 

There has been additionally been some research done on on-the-fly adaptation to completely unknown agents \cite{zand2022ontheflystrategyadaptationadhoc, fuchs2021theoryminddeepreinforcement, liu2023learningplaytrajectorygames, NEURIPS2020_65cf25ef}. These have found strong results, being able to learn the behaviors of partner and opposing agents over many iterations of play, but on-the-fly adaptation is a less demanding problem than ZSC, where you are expected to learn the best policy over multiple evaluation episodes instead of being able to cooperate well immediately. 

\begin{figure}[tbp]
    \centering
    \includegraphics[width=0.197\textwidth]{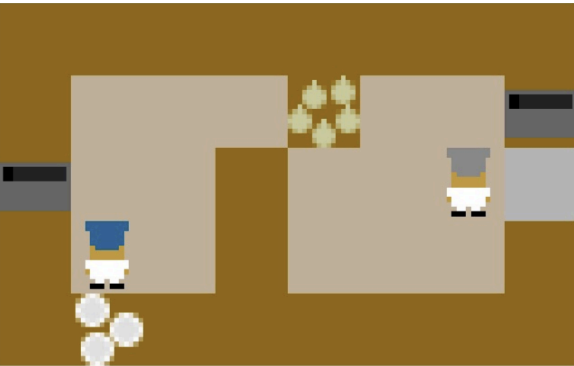}
    \includegraphics[width=0.2\textwidth]{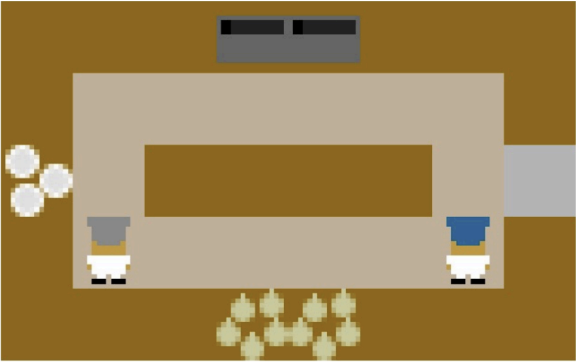} \\
    \vspace{6pt}\includegraphics[width=0.2\textwidth]{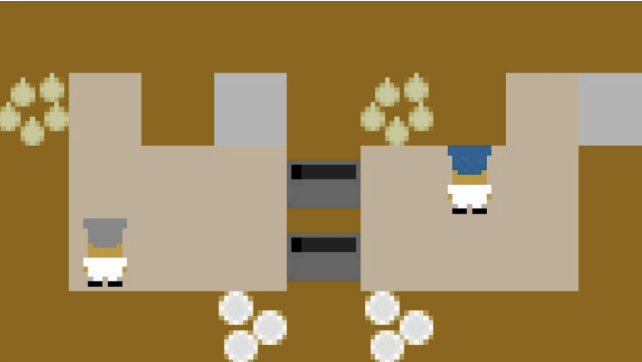}
    \caption{Three Overcooked Environments. Pictured in top left is Random0\_Medium, top right is Random3, and bottom is Unident\_S. Included are onions, pots, plates, stoves, and a delivery station. How would the blue and gray agents cooperate to the best extent possible without pre-coordination or knowing each other's strategies? How would we identify good performances across all 3 environments, including where strong cooperation is forced, like in Random0\_Medium or Unident\_S?}
    \label{fig:overcooked}
\end{figure}

We address this gap in considered policies by training ZSC agents with diverse reward shapings. We propose 4 novel methods of selecting these diverse sets of reward shapings (LLM-Based, Surrogate Network, Stratified Grid, and Random) and create ensemble models from populations trained on different shapings. Unlike existing ZSC methods that create diverse policies within a single reward function or randomly alter it slightly, this allows our trained agents to more accurately adapt to other agents trained  with different reward shapings by creating diversity across both policy space and reward functions. We evaluate our method on the Overcooked \cite{carroll2020utilitylearninghumanshumanai} environment, where results show 62.2\%-119.2\% improvement in sparse reward for the best selection methods compared to the baseline algorithms when cooperating with agents trained with unknown reward shapings.

Our contributions are as follows:
\begin{itemize}
    \item We introduce 4 shaping selection methods and an ensemble framework for combining ZSC models.
    \item We demonstrate that Stratified Grid sampling and MLP-based selection achieve the best generalization to unknown partners with unknown reward shapings.
\end{itemize}

\section{Problem Definition}
We formulate the ZSC problem in terms of a decentralized partially observable Markov decision process (Dec-POMDP). A Dec-POMDP $\mathbb{M}$ is defined as the tuple \(\mathbb{M} = (\mathbb{N}, \mathbb{S}, \mathbb{A}, \mathbb{T}, R, \mathbb{O}, \gamma, \mu)\), where \(\mathbb{N}\) is the set of agents, representing $N = |\mathbb{N}|$ cooperative agents, $\mathbb{S}$ is the joint state space shared by each agent, $\mathbb{A} = \prod_{k \in \mathbb{N}}a_k$ is the joint action space, where $a_k$ is the specific action space for agent $k$, $\mathbb{T}: \mathbb{S} \times \mathbb{A} \rightarrow \Delta \mathbb{S}$ is the transition function which determines the distribution of the next state given the current joint state and action, $R: \mathbb{S} \times \mathbb{A} \rightarrow \mathbb{R}$ is the reward function, $\mathbb{O}$ is the observation function, $\gamma$ is the discount factor, and $\mu$ is the initial state distribution. 

The objective of a policy $\pi$ in a Dec-POMDP is to maximize the expected return over finite time horizon $H$:
\[    
\pi^* = \text{argmax}_{\pi \in \Pi}J(\pi)
\]
where 
\[
J(\pi) = \mathbb{E}_\pi\left[\sum_{t=0}^H\gamma^tR(s_t, a_t)\bigg | \pi \right]
\]
where $\Pi = \{\pi = (\pi_1...\pi_N) | \pi^i = o^i_t \rightarrow a_i^t\}$ is the set of decentralized policies, with each individual policies selecting action $a^i$ based on their local observation $o^i$. 

The goal for ZSC is to learn a joint policy that generalizes well during cooperation with unknown agents and unknown policies, trained with different methods. Unlike traditional training and self-play algorithms, agents will not know each others policies during training or testing, meaning that conventions and pre-coordination attempts will fail. This gives rise to a specific performance metric for ZSC, the average cross-play (XP) loss, which for the two player case is given as:
\[
J_{XP}(\pi_1, \pi_2) = \frac 12(J(\pi^1_1, \pi^2_2) + J(\pi^1_2, \pi_1^2))
\]
where $\pi^1$ and $\pi^2$ represent the individual components of the policies (i.e. which player they are playing as).  

\section{Selecting Diverse Reward Shapings}

We aim to improve the generalization of ZSC and expand it to a wider context by considering the problem of both diverse reward shapings and diverse policies. We introduce methods for selecting reward shapings that are then used to generate a population of models during training, and finally evaluated together as an ensemble model with unknown partner agents that have identical sparse objectives but unknown reward shapings.

\subsection{Selection and Usage}
For our purposes, we consider a reward shaping as a set of weights, each referring to a specific environmental feature. We use the following notation for reward shapings and our selection methods: $\text{sparse}(s_t, a_t)$ is the sparse reward in state $s_t$ with action $a_t$, $f_k$ is a specific environmental feature used for reward shaping, $w_k$ is the weight applied to feature $f_k$, $P_i = (w_1; ...;w_k)$ is the reward shaping, a vector of weights applied to the features,  $F = |P_i|$ is the number of features used, and $P$ is the number of reward shapings generated by the selection method. This gives us a notation for the densification of a reward as 
\[
R(s_t, a_t) = \text{sparse}(s_t, a_t) + \sum_{k=1}^{F}w_kf_k
\]
The notion of densifying the reward with a weighted combination of features is the most common way of reward shaping, and is inbuilt into many MARL environments. 
We create and evaluate 4 methods of selecting reward shapings:

\textbf{LLM-Based:} Inspired by \cite{ma2024eurekahumanlevelrewarddesign}, we utilize LLMs to generate reward shapings. Specifically, we prompt Claude Sonnet 4.5 with a list of randomly generated $P_1...P_n$ ($n$ is arbitrary, we used 41 examples) as well as their individual training and testing results in terms of both sparse and shaped reward. We also supplied the code for both the environment and the agent, and asked it to generate a new set $P$ that would both perform well and result in an ending population of diverse policies and agents. Exact prompt and example output is given in Table \ref{tab:llm_prompt}. 

\textbf{Surrogate Network:} We train a Multi-Layer Perceptron (MLP) using supervised learning on the same examples given to the LLM. Specifically, for model $g(\theta)$ we input reward shaping $P_i$ and expect output total sparse reward $\text{sparse}_i$. This model then evaluates 1000 randomly generated shapings and then we select the best set of predicted results. Exact details on architecture are given in \ref{sec:surrogate_network}. 

\textbf{Stratified Grid:} We use Latin Hypercube Sampling \cite{ef76b040-2f28-37ba-b0c4-02ed99573416} to select a group of reward shapings. This works as a form of stratified sampling on each feature present. We divide the range of possible weights for each feature $f_k$ into $P$ strata of equal probability $1/P$ and sample once from each stratum. Calling that sample $X_k$, and performing the same process for each feature gives us an array of weights $w_{k, j}$, where $k$ is the associated feature. We then form random $P_i$ by selecting one unused weight from each column. This ensures that all possible weights are evenly represented and gives us a group of $P_i$ that reasonably represents the entire grid of possible reward shapings. 

\textbf{Random:} We uniformly randomly select each weight $w_k$ from the possible range for feature $f_k$ to form $P_i$. This gives us a reasonably diverse set of reward shapings, but without any guarantees or knowledge of how they might perform or whether they accurately represent the space.

\subsection{Policy Population Generation}
To actually implement these selection methods, we build off of the Trajectory Diversity (TrajeDi) method \cite{pmlr-v139-lupu21a}, which trains a population of models with diverse policies by implementing a Jensen-Shannon divergence term within the reward function. Specifically, TrajeDi trains a population of $n$ agents as well as an $(n+1)$-th agent to act as a Best Response (BR) to each other agent in the population, which is used as the final output of the training. The loss for TrajeDi is formulated as 
\begin{align*}
    \mathbb{L}(BR, \pi_1, ..., \pi_n) = -\bigg[&J(BR) + \alpha JSD_\gamma (\pi_1, ..., \pi_n)\\ &+ \sum_{i=1}^{n}(J_{XP}(BR, \pi_i) + J(\pi_i) )\bigg]
\end{align*}
where $JSD(\pi_1, ..., \pi_n)$  is the Jensen-Shannon divergence \cite{61115}. 

To generate our final agents, we use each selection method to generate $P$ reward shapings. We then train a population based on each of those unique shapings with the TrajeDi algorithm, and then combine these populations by using each BR agent as a single component of a final ensemble model, with the action chosen by the ensemble model being the mode action:
\[
\pi_{\text{ensemble}}(o_i) = \text{argmax}_{a_i \in A}\sum_{i=1}^N1[BR_i(o_i) = a_i]
\]
where $1[]$ is the indicator function. 

To compare this formulation with the baseline TrajeDi algorithm, we also compare an ensembled formulation of the baseline, equivalent to doing the same process as above but selecting the same base reward shaping each time. 

We compare both the original baseline, the ensembled baseline, and our selection methods by evaluating them in cooperation with agents that have been trained using randomly selected unknown reward shapings. These other agents are trained simply using MAPPO \cite{yu2022surprisingeffectivenessppocooperative}, a baseline MARL algorithm not made for ZSC. We also compare all these algorithms with Hidden-Utility Self-Play (HSP) \cite{yu2023learningzeroshotcooperationhumans}, explained further in the Related Work.

\section{Experimental Setup}
We conduct experiments on the Overcooked \cite{carroll2020utilitylearninghumanshumanai} environment. This is based on the cooking game of the same name, a cooperative game where player are given tasks to cook various dishes within a short time limit, having to perform various tasks such as cooking, chopping, and combining different ingredients. The goal in Overcooked is to deliver as many orders as possible, with delivering orders ahead of time giving you bonus score, and failing to deliver orders on time applying penalties. Overcooked has become one of the most popular environments for evaluating the coordination and cooperation of MARL agents due to the diversity of tasks and ease of use \cite{wang2024zscevalevaluationtoolkitbenchmark, 10.5555/3666122.3666241}. We utilize three specific environments within Overcooked - Random0\_Medium, Random3, and Unident\_S, all pictured in Figure \ref{fig:overcooked}. We select these three due to their range of difficulty - Random3 has all agents in the same layout, with no forced strong cooperation, while Random0\_Medium and Unident\_S either strongly encourage or force cooperation, with distinct separated areas and asymmetric resources. 

We use the code from ZSC-Eval \cite{wang2024zscevalevaluationtoolkitbenchmark}, which is a pre-built library for training and evaluating ZSC algorithms, including existing code for both the Overcooked environment and TrajeDi. We use $P=10$ reward shapings and train our agents with 100 million timesteps and then evaluate them over 10 seeds, each with 40 rollouts that each take 40000 timesteps. During training, we also perform a short evaluation of the models every 20 million timesteps. Note that these short evaluations during training are with agents that have the same reward shapings. 

We consider each reward shaping element has having a possible weight from 0 to 10. The reward shaping elements we used are as follows: PLACEMENT\_IN\_POT\_REW, which rewards the agent for placing an ingredient in a pot, DISH\_PICKUP\_REWARD, which rewards the agent for picking up a dish, SOUP\_PICKUP\_REWARD, which rewards the agent for picking up soup, DISH\_DISP\_DISTANCE\_REW, which rewards the agent for being close to the dish dispenser, POT\_DISTANCE\_REW, which rewards the agent for being close to a pot, and SOUP\_DISTANCE\_REW, which rewards the agent for being close to soup. 

We aim to answer three questions: Does introducing diverse reward shapings improve the performance over the baseline algorithm? Which selection methods produce the most diverse reward shapings? Is diversity of the reward shapings correlated with performance? 

\section{Results}
\begin{figure*}[htbp]
    \centering
    \includegraphics[width=\textwidth]{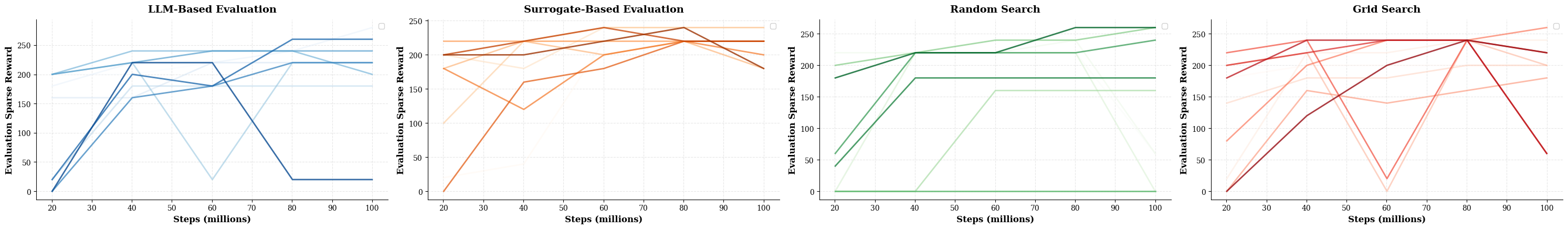}
    \caption{Evaluation runs during training. Each line shows the performance of a BR agent in one of the populations made by the selection method. Displayed is the sparse reward, evaluated once every 20 million timesteps during the 100 million training timesteps. }
    \label{fig:method_comparison}
\end{figure*}

\begin{table}[htbp]
\centering
\small
\renewcommand{\arraystretch}{1.5}
\begin{tabular}{l|c|c}
\hline
\textbf{Algorithm} & \textbf{Sparse Reward} & \textbf{vs Base. (ens.)} \\
\hline
LLM-Based & 104.1 $\pm$ 1.58 & +54.3\% \\
Surrogate Network & 123.2 $\pm$ 1.24 & +82.7\% \\
Stratified Grid & \textbf{131.4 $\pm$ 1.82} & +94.9\% \\
Random & 85.6 $\pm$ 1.69 & +27.0\% \\
\hline
HSP \cite{yu2023learningzeroshotcooperationhumans} & 26.8 $\pm$ 3.17 & -60.2\% \\
Baseline \cite{pmlr-v139-lupu21a} & \underline{20.6 $\pm$ 6.21} & -69.5\% \\
Baseline (ensembled) & 67.4 $\pm$ 0.64 & --- \\
\hline
\end{tabular}
\caption{Sparse Reward comparison during evaluation on Random3 environment. Values show mean $\pm$ 90\% CI.}
\label{tab:sparse_reward_random3}
\end{table}

\begin{table}[htbp]
\centering
\small
\renewcommand{\arraystretch}{1.5}
\begin{tabular}{l|c|c}
\hline
\textbf{Algorithm} & \textbf{Shaped Reward} & \textbf{vs Base. (ens.)} \\
\hline
LLM-Based & 90.1 $\pm$ 1.10 & +61.1\% \\
Surrogate Network & 103.0 $\pm$ 0.97 & +84.2\% \\
Stratified Grid & \textbf{107.4 $\pm$ 1.08} & +92.2\% \\
Random & 72.5 $\pm$ 1.10 & +29.6\% \\
\hline
HSP \cite{yu2023learningzeroshotcooperationhumans} & 24.4 $\pm$ 2.65 & -56.6\% \\
Baseline \cite{pmlr-v139-lupu21a} & \underline{20.9 $\pm$ 4.78} & -62.6\% \\
Baseline (ensembled) & 55.9 $\pm$ 0.39 & --- \\
\hline
\end{tabular}
\caption{Shaped Reward comparison during evaluation on Random3 environment. Values show mean $\pm$ 90\% CI.}
\label{tab:shaped_reward_random3}
\end{table}

\begin{table}[htbp]
\centering
\small
\renewcommand{\arraystretch}{1.5}
\begin{tabular}{l|c|c}
\hline
\textbf{Algorithm} & \textbf{Sparse Reward} & \textbf{vs Base. (ens.)} \\
\hline
LLM-Based & 47.1 $\pm$ 0.78 & +30.1\% \\
Surrogate Network & \textbf{59.3 $\pm$ 0.86} & +63.8\% \\
Stratified Grid & 58.7 $\pm$ 0.36 & +62.2\% \\
Random & 49.3 $\pm$ 0.74 & +36.2\% \\
\hline
HSP \cite{yu2023learningzeroshotcooperationhumans} & \underline{14.8 $\pm$ 1.53} & -59.2\% \\
Baseline \cite{pmlr-v139-lupu21a} & 15.0 $\pm$ 3.55 & -58.5\% \\
Baseline (ensembled) & 36.2 $\pm$ 0.60 & --- \\
\hline
\end{tabular}
\caption{Sparse Reward comparison on Random0\_Medium environment. Values show mean $\pm$ 90\% CI.}
\label{tab:sparse_reward_random0_medium}
\end{table}

\begin{table}[htbp]
\centering
\small
\renewcommand{\arraystretch}{1.5}
\begin{tabular}{l|c|c}
\hline
\textbf{Algorithm} & \textbf{Shaped Reward} & \textbf{vs Base. (ens.)} \\
\hline
LLM-Based & \textbf{146.1 $\pm$ 0.66} & +100.0\% \\
Surrogate Network & 141.5 $\pm$ 0.48 & +93.6\% \\
Stratified Grid & 136.0 $\pm$ 0.33 & +86.0\% \\
Random & 94.2 $\pm$ 0.46 & +29.0\% \\
\hline
HSP \cite{yu2023learningzeroshotcooperationhumans} & 38.8 $\pm$ 6.71 & -46.9\% \\
Baseline \cite{pmlr-v139-lupu21a} & \underline{32.6 $\pm$ 8.40} & -55.4\% \\
Baseline (ensembled) & 73.1 $\pm$ 0.30 & --- \\
\hline
\end{tabular}
\caption{Shaped Reward comparison on Random0\_Medium environment. Values show mean $\pm$ 90\% CI.}
\label{tab:shaped_reward_random0_medium}
\end{table}

\begin{table}[htbp]
\centering
\small
\renewcommand{\arraystretch}{1.5}
\begin{tabular}{l|c|c}
\hline
\textbf{Algorithm} & \textbf{Sparse Reward} & \textbf{vs Base. (ens.)} \\
\hline
LLM-Based & 72.5 $\pm$ 0.99 & +102.5\% \\
Surrogate Network & \textbf{78.5 $\pm$ 1.04} & +119.2\% \\
Stratified Grid & 69.7 $\pm$ 0.88 & +94.7\% \\
Random & 54.2 $\pm$ 1.09 & +51.4\% \\
\hline
HSP \cite{yu2023learningzeroshotcooperationhumans} & 24.9 $\pm$ 6.39 & -30.4\% \\
Baseline \cite{pmlr-v139-lupu21a} & \underline{22.5 $\pm$ 8.59} & -37.2\% \\
Baseline (ensembled) & 35.8 $\pm$ 0.61 & --- \\
\hline
\end{tabular}
\caption{Sparse Reward comparison on Unident\_s environment. Values show mean $\pm$ 90\% CI.}
\label{tab:sparse_reward_unident_s}
\end{table}

\begin{table}[htbp]
\centering
\small
\renewcommand{\arraystretch}{1.5}
\begin{tabular}{l|c|c}
\hline
\textbf{Algorithm} & \textbf{Shaped Reward} & \textbf{vs Base. (ens.)} \\
\hline
LLM-Based & 56.9 $\pm$ 0.65 & +44.1\% \\
Surrogate Network & \textbf{67.8 $\pm$ 0.56} & +71.6\% \\
Stratified Grid & 66.6 $\pm$ 0.82 & +68.6\% \\
Random & 51.7 $\pm$ 0.61 & +30.9\% \\
\hline
HSP \cite{yu2023learningzeroshotcooperationhumans} & \underline{20.4 $\pm$ 4.45} & -48.4\% \\
Baseline \cite{pmlr-v139-lupu21a} & 24.2 $\pm$ 9.7 & -39.7\% \\
Baseline (ensembled) & 39.5 $\pm$ 0.83 & --- \\
\hline
\end{tabular}
\caption{Shaped Reward comparison on Unident\_s environment. Values show mean $\pm$ 90\% CI.}
\label{tab:shaped_reward_unidient_s}
\end{table}

\subsection{Performance}
Seen in Table \ref{tab:sparse_reward_random3} through Table\ref{tab:shaped_reward_unidient_s}, our algorithms improve upon the baseline in all cases and environments. Specifically, Surrogate Network and Stratified Grid exhibit the most significant improvements, seeing anywhere from a $\sim$60\% improvement to a $\sim$120\% improvement across sparse and shaped rewards, and almost always performing the best across all methods. We can see that a significant improvement is also obtained from the ensemble nature of the methods, with the ensembled version of TrajeDi generally exhibiting an 40-70\% improvement over the non-ensembled version. This is likely partially due to the fact that these evaluations were performed against models using unknown reward shapings and the more robust, adaptable, and resilient properties of ensemble models. However, we can clearly see that not all improvement is exhibited to that, with the ensembled Baseline always being improved by our methods with at least 27\% improvement. The poor performance of random selection vs. all other methods is likely due to the fact that no guarantees or attempts were made on random to either improve the performance or diversity like with all other methods, but even so, it improves on the baselines by a significant amount. 

\begin{table}[htbp]
\centering
\small
\renewcommand{\arraystretch}{1.5}
\begin{tabular}{l|c|c}
\hline
\textbf{Selection Method} & \textbf{Avg. Stdev} & \textbf{\% Range Covered}\\
\hline
LLM-Based & $3.0558$ & $83.3$\\
Surrogate Network & $2.2926$ & $63.3$ \\
Stratified Grid & $2.9452$ & $96.7$ \\
Random & $3.2352$ & $91.7$ \\
\hline
\end{tabular}
\caption{Diversity metrics for each selection method. Averaged Standard Deviation is the average of the standard deviation of each weight across the group of reward shapings. \% Range covered is the average of range covered (within the possible 0-10) by the values associated with each weight.}
\label{tab:weights_stdev}
\end{table}

\subsection{Diversity}
Shown in Table \ref{tab:weights_stdev} (and with more specificity in Table \ref{tab:shaping_params_specific}) we can see the averaged standard deviation across the produced weights for each selection method. That is, taking the standard deviation for the values given to each weight, and then averaging those values across all the weights. These values are then averaged across all 3 environments to display in the table above. Notable is that even though the Surrogate Network selection maintains a significant standard deviation for values 0-10, it is significantly lower than all other selection methods, by around $0.65$. All three other selection methods hover around a much closer range, from about $2.95-3.25$. One thing of note is that performance increases as standard deviation decreases between Stratified Grid, LLM-Based, and Random selection, but Surrogate Network selection performance decreases even as the standard deviation decreases. It's likely that the most optimal variance lays between the approaches of Surrogate Network (most closely optimal and based on existing data) and Stratified Grid (ignoring existing data and giving the most varied data reasonably possible).  

Standard deviation is not the only method of diversity - we can also see in Table \ref{tab:weights_stdev} that Stratified Grid covers the largest range of each weight on average, slightly higher than random selection, even though Random selection has a higher standard deviation. This is due to the fact that Random Selection can easily select multiple reward shapings that are very similar to each other, creating results that clump and do not accurately reflect across the entire space. Surrogate Network performs excellently despite it's low coverage due to picking rewards that have achieved the highest sparse reward in the past, as with 1000 rewards to choose from it is likely to consistently pick rewards most similar to those that have performed excellently, and even if those are not especially diverse, their individual performances and reasonable coverage can likely make up for it. Even though both methods were given the exact same data, Surrogate Network likely picks a much smaller range of values compared to LLM-Based due to the LLM being prompted specifically to choose a set of shapings that would result in diverse policies. 

Seen in Figure \ref{fig:method_comparison} is the evaluation steps run during training time on the Random3 environment, showing the performance of each BR agent for each reward shaping selected by the selection method. Surrogate Network converges to a small set of excellently performing policies, showing both its lack of diversity and its weight on individual performance due to how it was trained and used. LLM-Based is similar, but with a slightly wider range, and 1 policy that performs badly in the ending two evaluation steps, likely due to the differences mentioned above. Stratified Grid shows the most evenly spaced out performance, with most reward shapings performing reasonably well, but not being concentrated in a specific area or shape like the model-based selections. Random has performances clumped into three areas, with 3 reward shapings demonstrating very poor performance over time, likely due to both it's high diversity but lack of constraints on representative diversity.

\section{Conclusion and Future Work}

\textbf{Introducing multiple reward shapings vastly improves performance:} Even with the most naive method of selection, we still find significant improvements over using the same reward shaping for each agent.

\textbf{Our methods can generate diverse reward shapings:} Our methods generated significantly more diverse methods than sticking closely to one reward shaping set, with even the least diverse method covering almost 2/3 of the space.

\textbf{Diversity is useful, but not the only factor of importance:} While the most successful method was Stratified Grid sampling, which is the most representative and reasonably diverse, the second most successful method was Surrogate Network selection, the least diverse and representative.

Future work will focus on expanding the breadth of testing, expanding to more environments and more baseline algorithms. In addition, modifying the training and prompting of the Surrogate Network and LLM may yield increased performance for those methods. Specifically, introducing a diversity metric to encourage the Surrogate Network to evaluate a more representative set of samples higher may improve performance. This may allow us to take the best aspects from both Surrogate Network and Stratified Grid sampling.

\bibliographystyle{unsrt}  
\bibliography{references}  

@InProceedings{pmlr-v139-lupu21a,
  title = 	 {Trajectory Diversity for Zero-Shot Coordination},
  author =       {Lupu, Andrei and Cui, Brandon and Hu, Hengyuan and Foerster, Jakob},
  booktitle = 	 {Proceedings of the 38th International Conference on Machine Learning},
  pages = 	 {7204--7213},
  year = 	 {2021},
  editor = 	 {Meila, Marina and Zhang, Tong},
  volume = 	 {139},
  series = 	 {Proceedings of Machine Learning Research},
  month = 	 {18--24 Jul},
  publisher =    {PMLR},
  url = 	 {https://proceedings.mlr.press/v139/lupu21a.html}
}

@misc{ma2024eurekahumanlevelrewarddesign,
      title={Eureka: Human-Level Reward Design via Coding Large Language Models}, 
      author={Yecheng Jason Ma and William Liang and Guanzhi Wang and De-An Huang and Osbert Bastani and Dinesh Jayaraman and Yuke Zhu and Linxi Fan and Anima Anandkumar},
      year={2024},
      eprint={2310.12931},
      archivePrefix={arXiv},
      primaryClass={cs.RO},
      url={https://arxiv.org/abs/2310.12931}, 
}

@InProceedings{pmlr-v9-glorot10a,
  title = 	 {Understanding the difficulty of training deep feedforward neural networks},
  author = 	 {Glorot, Xavier and Bengio, Yoshua},
  booktitle = 	 {Proceedings of the Thirteenth International Conference on Artificial Intelligence and Statistics},
  pages = 	 {249--256},
  year = 	 {2010},
  editor = 	 {Teh, Yee Whye and Titterington, Mike},
  volume = 	 {9},
  series = 	 {Proceedings of Machine Learning Research},
  address = 	 {Chia Laguna Resort, Sardinia, Italy},
  month = 	 {13--15 May},
  publisher =    {PMLR},
  url = 	 {https://proceedings.mlr.press/v9/glorot10a.html}
}

@article{ef76b040-2f28-37ba-b0c4-02ed99573416,
 ISSN = {00401706},
 URL = {http://www.jstor.org/stable/1268522},
 author = {M. D. McKay and R. J. Beckman and W. J. Conover},
 journal = {Technometrics},
 number = {2},
 pages = {239--245},
 publisher = {[Taylor & Francis, Ltd., American Statistical Association, American Society for Quality]},
 title = {A Comparison of Three Methods for Selecting Values of Input Variables in the Analysis of Output from a Computer Code},
 urldate = {2025-12-07},
 volume = {21},
 year = {1979}
}

@article{10.1162/neco.1994.6.2.215,
author = {Tesauro, Gerald},
title = {TD-Gammon, a self-teaching backgammon program, achieves master-level play},
year = {1994},
issue_date = {March 1994},
publisher = {MIT Press},
address = {Cambridge, MA, USA},
volume = {6},
number = {2},
issn = {0899-7667},
url = {https://doi.org/10.1162/neco.1994.6.2.215},
doi = {10.1162/neco.1994.6.2.215},
journal = {Neural Comput.},
month = mar,
pages = {215–219},
numpages = {5}
}

@inproceedings{NEURIPS2020_65cf25ef,
 author = {Sessa, Pier Giuseppe and Bogunovic, Ilija and Kamgarpour, Maryam and Krause, Andreas},
 booktitle = {Advances in Neural Information Processing Systems},
 editor = {H. Larochelle and M. Ranzato and R. Hadsell and M.F. Balcan and H. Lin},
 pages = {8971--8981},
 publisher = {Curran Associates, Inc.},
 title = {Learning to Play Sequential Games versus Unknown Opponents},
 volume = {33},
 year = {2020}
}

@misc{liu2023learningplaytrajectorygames,
      title={Learning to Play Trajectory Games Against Opponents with Unknown Objectives}, 
      author={Xinjie Liu and Lasse Peters and Javier Alonso-Mora},
      year={2023},
      eprint={2211.13779},
      archivePrefix={arXiv},
      primaryClass={cs.RO},
      url={https://arxiv.org/abs/2211.13779}, 
}

@misc{foerster2019bayesianactiondecoderdeep,
      title={Bayesian Action Decoder for Deep Multi-Agent Reinforcement Learning}, 
      author={Jakob N. Foerster and Francis Song and Edward Hughes and Neil Burch and Iain Dunning and Shimon Whiteson and Matthew Botvinick and Michael Bowling},
      year={2019},
      eprint={1811.01458},
      archivePrefix={arXiv},
      primaryClass={cs.MA},
      url={https://arxiv.org/abs/1811.01458}, 
}

@misc{li2025atdronebenchmarkingadaptiveteaming,
      title={AT-Drone: Benchmarking Adaptive Teaming in Multi-Drone Pursuit}, 
      author={Yang Li and Junfan Chen and Feng Xue and Jiabin Qiu and Wenbin Li and Qingrui Zhang and Ying Wen and Wei Pan},
      year={2025},
      eprint={2502.09762},
      archivePrefix={arXiv},
      primaryClass={cs.RO},
      url={https://arxiv.org/abs/2502.09762}, 
}

@misc{hu2021otherplayzeroshotcoordination,
      title={"Other-Play" for Zero-Shot Coordination}, 
      author={Hengyuan Hu and Adam Lerer and Alex Peysakhovich and Jakob Foerster},
      year={2021},
      eprint={2003.02979},
      archivePrefix={arXiv},
      primaryClass={cs.AI},
      url={https://arxiv.org/abs/2003.02979}, 
}

@misc{xue2024heterogeneousmultiagentzeroshotcoordination,
      title={Heterogeneous Multi-agent Zero-Shot Coordination by Coevolution}, 
      author={Ke Xue and Yutong Wang and Cong Guan and Lei Yuan and Haobo Fu and Qiang Fu and Chao Qian and Yang Yu},
      year={2024},
      eprint={2208.04957},
      archivePrefix={arXiv},
      primaryClass={cs.NE},
      url={https://arxiv.org/abs/2208.04957}, 
}

@misc{ibrahim2024comprehensiveoverviewrewardengineering,
      title={Comprehensive Overview of Reward Engineering and Shaping in Advancing Reinforcement Learning Applications}, 
      author={Sinan Ibrahim and Mostafa Mostafa and Ali Jnadi and Hadi Salloum and Pavel Osinenko},
      year={2024},
      eprint={2408.10215},
      archivePrefix={arXiv},
      primaryClass={cs.LG},
      url={https://arxiv.org/abs/2408.10215}, 
}

@misc{carroll2020utilitylearninghumanshumanai,
      title={On the Utility of Learning about Humans for Human-AI Coordination}, 
      author={Micah Carroll and Rohin Shah and Mark K. Ho and Thomas L. Griffiths and Sanjit A. Seshia and Pieter Abbeel and Anca Dragan},
      year={2020},
      eprint={1910.05789},
      archivePrefix={arXiv},
      primaryClass={cs.LG},
      url={https://arxiv.org/abs/1910.05789}, 
}

@misc{wang2024zscevalevaluationtoolkitbenchmark,
      title={ZSC-Eval: An Evaluation Toolkit and Benchmark for Multi-agent Zero-shot Coordination}, 
      author={Xihuai Wang and Shao Zhang and Wenhao Zhang and Wentao Dong and Jingxiao Chen and Ying Wen and Weinan Zhang},
      year={2024},
      eprint={2310.05208},
      archivePrefix={arXiv},
      primaryClass={cs.AI},
      url={https://arxiv.org/abs/2310.05208}, 
}

@misc{yu2022surprisingeffectivenessppocooperative,
      title={The Surprising Effectiveness of PPO in Cooperative, Multi-Agent Games}, 
      author={Chao Yu and Akash Velu and Eugene Vinitsky and Jiaxuan Gao and Yu Wang and Alexandre Bayen and Yi Wu},
      year={2022},
      eprint={2103.01955},
      archivePrefix={arXiv},
      primaryClass={cs.LG},
      url={https://arxiv.org/abs/2103.01955}, 
}

@misc{zand2022ontheflystrategyadaptationadhoc,
      title={On-the-fly Strategy Adaptation for ad-hoc Agent Coordination}, 
      author={Jaleh Zand and Jack Parker-Holder and Stephen J. Roberts},
      year={2022},
      eprint={2203.08015},
      archivePrefix={arXiv},
      primaryClass={cs.LG},
      url={https://arxiv.org/abs/2203.08015}, 
}

@misc{fuchs2021theoryminddeepreinforcement,
      title={Theory of Mind for Deep Reinforcement Learning in Hanabi}, 
      author={Andrew Fuchs and Michael Walton and Theresa Chadwick and Doug Lange},
      year={2021},
      eprint={2101.09328},
      archivePrefix={arXiv},
      primaryClass={cs.AI},
      url={https://arxiv.org/abs/2101.09328}, 
}

@inproceedings{10.5555/3666122.3666241,
author = {Yan, Xue and Guo, Jiaxian and Lou, Xingzhou and Wang, Jun and Zhang, Haifeng and Du, Yali},
title = {An efficient end-to-end training approach for zero-shot human-AI coordination},
year = {2023},
publisher = {Curran Associates Inc.},
address = {Red Hook, NY, USA},
booktitle = {Proceedings of the 37th International Conference on Neural Information Processing Systems},
articleno = {119},
numpages = {23},
location = {New Orleans, LA, USA},
series = {NIPS '23}
}

@misc{yu2023learningzeroshotcooperationhumans,
      title={Learning Zero-Shot Cooperation with Humans, Assuming Humans Are Biased}, 
      author={Chao Yu and Jiaxuan Gao and Weilin Liu and Botian Xu and Hao Tang and Jiaqi Yang and Yu Wang and Yi Wu},
      year={2023},
      eprint={2302.01605},
      archivePrefix={arXiv},
      primaryClass={cs.AI},
      url={https://arxiv.org/abs/2302.01605}, 
}

@ARTICLE{61115,
  author={Lin, J.},
  journal={IEEE Transactions on Information Theory}, 
  title={Divergence measures based on the Shannon entropy}, 
  year={1991},
  volume={37},
  number={1},
  pages={145-151},
  doi={10.1109/18.61115}}

\appendix
\section{Surrogate Network Architecture}
\label{sec:surrogate_network}
Due to the small amount of examples (just 41), we utilized a small model (6 inputs with two hidden layers each with 32 and 16 units), as well as 30\% dropout and Xavier initialization \cite{pmlr-v9-glorot10a}. We train for 500 epochs with 0.001 learning rate and a batch size of 8 to prevent overfitting. 

\section{MARL Hyperparameters}
We use the default Hyperparameters given used for ZSC-Eval \cite{wang2024zscevalevaluationtoolkitbenchmark}. This includes a $\gamma$ of $0.99$, learning rate of $0.0005$, and clip\_param of $0.2$. 

\section{Qualitative Analysis for LLM-Based and Surrogate Network}

\begin{table*}[htbp]
\centering
\renewcommand{\arraystretch}{1.5}
\begin{tabular}{l|cc|cc|cc|cc}
\hline
\textbf{Parameter} & \multicolumn{2}{c|}{\textbf{LLM-Based}} & \multicolumn{2}{c|}{\textbf{Surrogate Network}} & \multicolumn{2}{c|}{\textbf{Stratified Grid}} & \multicolumn{2}{c}{\textbf{Random}} \\
 & \textbf{Mean} & \textbf{Stdev} & \textbf{Mean} & \textbf{Stdev} & \textbf{Mean} & \textbf{Stdev} & \textbf{Mean} & \textbf{Stdev} \\
\hline
PLACEMENT\_IN\_POT\_REW & $7.7$ & $2.00$ & $9.6$ & $1.29$ & $3.5$ & $3.21$ & $4.9$ & $3.31$ \\
DISH\_PICKUP\_REWARD & $6.0$ & $3.80$ & $9.5$ & $1.65$ & $5.3$ & $3.13$ & $4.8$ & $3.79$ \\
SOUP\_PICKUP\_REWARD & $4.4$ & $2.95$ & $9.7$ & $2.00$ & $5.0$ & $2.87$ & $4.5$ & $2.95$ \\
DISH\_DISP\_DISTANCE\_REW & $3.9$ & $2.96$ & $5.0$ & $3.86$ & $3.8$ & $2.78$ & $5.0$ & $3.30$ \\
POT\_DISTANCE\_REW & $4.1$ & $3.03$ & $1.3$ & $1.83$ & $5.9$ & $3.14$ & $4.0$ & $3.68$ \\
SOUP\_DISTANCE\_REW & $3.8$ & $3.58$ & $6.7$ & $3.13$ & $5.2$ & $4.29$ & $7.6$ & $2.37$ \\
\hline
\textbf{Average} & $\mathbf{5.0}$ & $\mathbf{3.06}$ & $\mathbf{7.0}$ & $\mathbf{2.29}$ & $\mathbf{4.8}$ & $\mathbf{3.24}$ & $\mathbf{5.1}$ & $\mathbf{3.24}$ \\
\hline
\end{tabular}
\caption{More specific mean and standard deviation for selection methods.}
\label{tab:shaping_params_specific}
\end{table*}

We can see a more detailed breakdown of variables in Table \ref{tab:shaping_params_specific} compared to the results in the main paper. Random has a couple outliers, but in general is very close to what you would expect, around 5.0 mean and around 3.0 standard deviation. Stratified Grid has a few more outliers, but still hovers around the 4-6 range with around 3.0 standard deviation. LLM-Based and Surrogate Network show more interesting results, with a lot of values (especially for Surrogate Network) abnormally high or with significantly lower standard deviations. This is likely due to the best performing reward shapings being ones that are strongly shaped, as that allows the model to densify the reward to a very detailed extent. 

\section{LLM Prompts}
Displayed in Table \ref{tab:llm_prompt} is the specific prompt used for the LLM to generate new reward shapings. You can see the specific emphasis on creating diverse shapings and how that played in to the LLMs rationale. You can also see how the LLM attempts to create specific strategies for each reward shaping cofiguration. 
\begin{table*}[htbp]
  \caption{The following table contains an example of the prompt used for the LLM and its response.}
  \label{tab:llm_prompt}
  \begin{mybody}
  \scriptsize{
    \textbf{System:} 
    Attached are two files. \\
    1: Part of the code for the Overcooked reinforcement learning environment, which has multiple possible shaped rewards. \\
    2: Training results from training populations of agents using specific reward shaping weights.\\
    
    I want you to propose to me, given the attached 2 files, a set of 10 reward shaping weights (using the same 6 shaped infos as the results file). I want you to prioritize two things. \\
    1: The diversity of your proposed reward shaping weights (not making them too similar to each other). You would prefer to propose a set of 10 weights that will create a set of 10 diverse policies. \\
    2: The estimated performances of your proposed weights. \\

Propose to me a format like this:
[{params...}, {params...}, ...] \\

\textbf{[Attached results file:]} \\

[\\
  {\\
    "folder": "traj-S1-s10-4-2-random", \\
    "pop\_num": 4, \\
    "seed": 2, \\
    "run\_type": "random", \\
    "best\_reward": 468.9725875854492, \\
    "reward\_shaping\_params": { \\
      "PLACEMENT\_IN\_POT\_REW": 10, \\
      "DISH\_PICKUP\_REWARD": 8, \\
      "SOUP\_PICKUP\_REWARD": 2, \\
      "DISH\_DISP\_DISTANCE\_REW": 3, \\
      "POT\_DISTANCE\_REW": 5, \\
      "SOUP\_DISTANCE\_REW": 0
    }\\
  },\\
  ...\\
  ]\\

\textbf{[Attached Overcooked Environment Code]} \\

\textbf{LLM:} \\
json[ \\
  \{ \\
    "PLACEMENT\_IN\_POT\_REW": 9, \\
    "DISH\_PICKUP\_REWARD": 10, \\
    "SOUP\_PICKUP\_REWARD": 7, \\
    "DISH\_DISP\_DISTANCE\_REW": 9, \\
    "POT\_DISTANCE\_REW": 0, \\
    "SOUP\_DISTANCE\_REW": 2 \\
 \}, \\
  ... \\
  ] \\
  Rationale: \\
High performers selected: I prioritized configurations from the top-performing runs (537, 505, 481, 475, 468, 464, 457, 454, 450 reward range) \\
Diverse behavioral strategies: \\
Configs 1, 3, 6: Emphasize early-game actions (placement + dish pickup) with varying distance rewards \\
Config 2, 7, 8: Focus heavily on pot placement with minimal soup delivery shaping \\
Config 4: Balanced approach with strong soup pickup emphasis \\
Config 5, 10: Low dish pickup, encouraging different coordination patterns \\
Config 9: Medium values across the board for stable coordination \\
Key diversity dimensions: \\
DISH\_PICKUP\_REWARD: ranges from 0 to 10 (creates agents with different item prioritization) \\
Distance rewards: varied combinations to encourage different movement patterns
PLACEMENT vs SOUP\_PICKUP trade-offs: different temporal focuses in the task pipeline \\

This set should produce diverse policies while maintaining strong performance potential. \\

    }
  \end{mybody}
\end{table*}

\end{document}